\documentclass{article} 
\usepackage[preprint]{colm2026}

\usepackage{microtype}
\usepackage{hyperref}
\usepackage{url}
\usepackage{booktabs}
\usepackage{xcolor}
\usepackage{colortbl}
\usepackage{enumitem}
\usepackage{lineno}
\usepackage{amsmath}
\usepackage{graphicx}
\usepackage{wrapfig}

\usepackage{tcolorbox}
\usepackage{amssymb}       
\usepackage{enumitem} 
\usepackage{booktabs}
\usepackage{caption}

\definecolor{darkblue}{rgb}{0, 0, 0.5}
\hypersetup{colorlinks=true, citecolor=darkblue, linkcolor=darkblue, urlcolor=darkblue}

\definecolor{rowgreen}{HTML}{F1F8F2}      
\definecolor{upgreen}{HTML}{2E7D32}       
\definecolor{downorg}{HTML}{D4A017}       

\newcommand{\oursrow}{\rowcolor{rowgreen}}
\newcommand{\inc}[1]{{\scriptsize\textcolor{upgreen}{$_{\uparrow#1}$}}}

\definecolor{phaseblue}{RGB}{30,100,180}
\newcommand{\phasecomment}[1]{\textcolor{phaseblue}{\textit{#1}}}
\newcommand{\sidecomment}[1]{\hfill \textcolor{gray}{\small$\triangleright$ #1}}

\title{ARISE: Agent Reasoning with Intrinsic Skill Evolution in Hierarchical Reinforcement Learning}

\author{
  Yu Li\textsuperscript{1}, Rui Miao\textsuperscript{2}, Zhengling Qi\textsuperscript{3}, Tian Lan\textsuperscript{1}\\
  \textsuperscript{1}Department of Electrical and Computer Engineering, George Washington University \\
  \textsuperscript{2}Department of Mathematical Sciences, University of Texas at Dallas\\
  \textsuperscript{3}School of Business,
  George Washington University \\
  \texttt{\{yul, tlan, qizhengling\}@gwu.edu}, \texttt{rui.miao@utdallas.edu}
}

\usepackage{amsthm}

\usepackage{algorithm}
\usepackage{algorithmic}
\usepackage{pifont}

\newcommand{\n}{ARISE}
\DeclareMathOperator*{\argmax}{arg\,max}

\begin{document}

\ifcolmsubmission
\linenumbers
\fi

\maketitle

\begin{abstract}
The dominant paradigm for improving mathematical reasoning in language models relies on Reinforcement Learning with verifiable rewards. Yet existing methods treat each problem instance in isolation without leveraging the reusable strategies that emerge and accumulate during training. To this end, we introduce ARISE (Agent Reasoning via Intrinsic Skill Evolution), a hierarchical reinforcement learning framework in which a shared policy operates both as a high-level Skills Manager and a low-level Worker. The Manager maintains a tiered skill library through two complementary mechanisms: before execution, a policy-driven selection mechanism retrieves relevant skills to condition future rollouts; after execution, a dedicated skill generation rollout performs structured summarization of successful solution traces. A hierarchical reward design guides the co-evolution of reasoning ability and library quality. Experiments on two base models and seven benchmarks spanning both competition mathematics and Omni-MATH show that ARISE consistently outperforms GRPO-family algorithms and memory-augmented baselines, with particularly notable gains on out-of-distribution tasks. Ablation studies confirm that each component contributes to the observed improvements and that library quality and reasoning performance improve in tandem throughout training.
\end{abstract}

\section{Introduction}

\begin{figure*}[hbt]
    \centering
    \includegraphics[width=\textwidth]{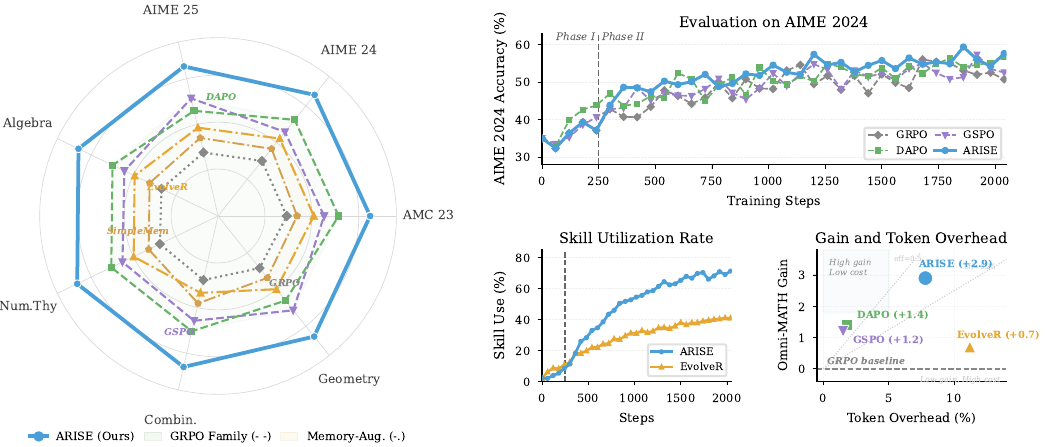}
    \caption{Overview of ARISE on Qwen3-4B, showing performance across seven benchmarks, training reward dynamics on DeepScaleR, skill utilization over training, and accuracy gain versus token overhead relative to GRPO.}
    \label{fig:intro}
\end{figure*}

Reinforcement learning with verifiable rewards has emerged as a compelling paradigm for training mathematical reasoning in large language models, enabling policies to improve through trial-and-error without relying on expensive human annotation~\citep{guo2025deepseek}. 
Despite strong performance on standard benchmarks, existing methods solve each problem instance via a separate process: Once a rollout concludes, the successful reasoning strategies generated in the process are discarded rather than retained and accumulated for future use~\citep{sun2025survey,zhang2025landscape}. 

A natural remedy is to equip the agent with a persistent skill library that accumulates reusable reasoning strategies over time~\citep{wang2025reinforcement,wei2026agentic}. 
Recent work has demonstrated the value of organizing past experience into structured skills and retrieving them at inference or training time~\citep{wu2025evolver, xia2026skillrl, wang2025reinforcement}, allowing agents to accumulate and transfer knowledge across problems and avoid redundant exploration. 
However, existing approaches share a fundamental limitation: skill management, including both skill selection before execution and skill summarization after execution, is delegated to an external retriever, preventing the policy gradient from directly shaping skill selection.
Furthermore, the skill library is updated independently of the RL objective, breaking the feedback loop between policy improvement and library enrichment~\citep{hao2024skill}.

We present~\n, an integrated, hierarchical reinforcement learning framework for Agent Reasoning via Intrinsic Skill Evolution, that addresses both limitations through a unified design. 
The key insight is that the skill library should not be a static external resource but an intrinsic component of the agent's state, co-evolving with the policy throughout training. 
\n realizes a Manager-Worker hierarchy within a single shared policy $\pi_\theta$: the manager selects skills using the policy's own log-probabilities and generates new skills by summarizing successful solution traces, while the worker generates solution traces conditioned on the selected skill. 
Since the same parameters govern both skill selection and solution generation, the advantage signal from the hierarchical reward propagates end-to-end, reinforcing selection preferences for skills that demonstrably improve reasoning outcomes.

To incentivize skill utilization, we introduce a hierarchical reward 
$R \in \{r_0, r_1, r_2\}$ with $r_2 > r_1 > r_0$, distinguishing 
correct solutions with skill augmentation ($r_2$), without ($r_1$), 
and incorrect solutions ($r_0$). 
Under group-relative advantage, this differential signal steers the 
policy toward consistently leveraging useful skills. 
The skill library adopts a two-tier cache-reservoir architecture with 
five management operations, maintaining a compact active pool while 
preserving skills that may regain relevance as training progresses. 
Training proceeds in two phases: a warm-up that builds the base policy 
and populates the library via $O_{G+1}$, followed by a skill-augmented 
phase activating the full hierarchical pipeline.

We evaluate \n\ on two instruction-tuned base models, Qwen3-4B-Instruct-2507~\citep{qwen3technicalreport} and Phi-4-mini-instruct~\citep{abouelenin2025phi}, trained on the DeepScaleR dataset. 
Results on both in-distribution competition benchmarks and the out-of-distribution benchmark with four mathematical domains~\citep{gao2024omni} demonstrate consistent improvements over GRPO-family baselines and existing memory- and skill-augmented methods.

Our main contributions are summarized as follows:
\begin{itemize}[leftmargin=1.2em]
  \item We propose the Evolving-Skill MDP, a formal framework that models the skill library as an endogenous component of the agent's state, enabling joint optimization of policy and library under a unified RL objective.
  \item We introduce a policy-driven skill selection mechanism based on conditional log-probability scoring, allowing the policy gradient to directly shape selection preferences end-to-end without relying on an external retriever.
  \item We design a hierarchical reward and a two-tier skill library architecture that together create a co-evolutionary dynamic between policy improvement and library enrichment.
  \item Empirical results on competition and Olympiad-level benchmarks demonstrate that \n\ consistently outperforms both vanilla GRPO variants and skill-augmented baselines across two base models.
\end{itemize}
\section{Backgrounds}
\label{sec:background}

Reinforcement learning has emerged as a powerful paradigm for enhancing the reasoning capabilities of large language models. Early RLHF methods~\citep{ouyang2022training} relied on learned reward models trained on human preferences, but are susceptible to reward model overoptimization~\citep{gao2023scaling} and require expensive annotation.
\citet{guo2025deepseek} introduced Group Relative Policy Optimization (GRPO), a critic-free variant of PPO~\citep{schulman2017proximal} that estimates advantages via normalized rewards within a group of $G$ rollouts per query, giving rise to the Reinforcement Learning with Verifiable Rewards (RLVR) paradigm now dominant in mathematical reasoning. 
Several refinements have since addressed GRPO's limitations: Dr.GRPO~\citep{liu2025understanding} corrects normalization biases in advantage estimation; DAPO~\citep{yu2025dapo} introduces asymmetric clipping and dynamic sampling to mitigate entropy collapse; and GSPO~\citep{zheng2025group} replaces per-token importance ratios with sequence-level correction to better align with the reward signal.

A complementary line of research augments LLM agents with external memory or reusable skill structures to enable experience transfer across episodes.
Inference-time methods such as Reflexion~\citep{shinn2023reflexion}, ExpeL~\citep{zhao2024expel}, and SimpleMem~\citep{liu2026simplemem} retrieve past trajectories or distilled knowledge into the agent's context, but the resulting memory is populated independently of policy learning and remains fixed once constructed. 
More recent work integrates skill structures directly into the training loop: EvolveR~\citep{wu2025evolver} maintains a co-evolving skill library and SkillRL~\citep{xia2026skillrl} builds a hierarchical skill bank via trajectory distillation. 
Building on this direction, SAGE~\citep{wang2025reinforcement} further incorporates skill augmentation into GRPO through sequential rollouts, coupling skill generation with policy optimization in a unified framework.
\section{\n: Hierarchical Agent RL with Evolving Skills}
\label{sec:method}

We present \n, a hierarchical reinforcement learning framework that addresses two limitations of existing skill-augmented approaches: skill selection is delegated to an external retriever decoupled from the policy gradient, and the skill library is updated independently of the RL objective.

\begin{figure}[!t]
    \centering
    \includegraphics[width=1.0\linewidth]{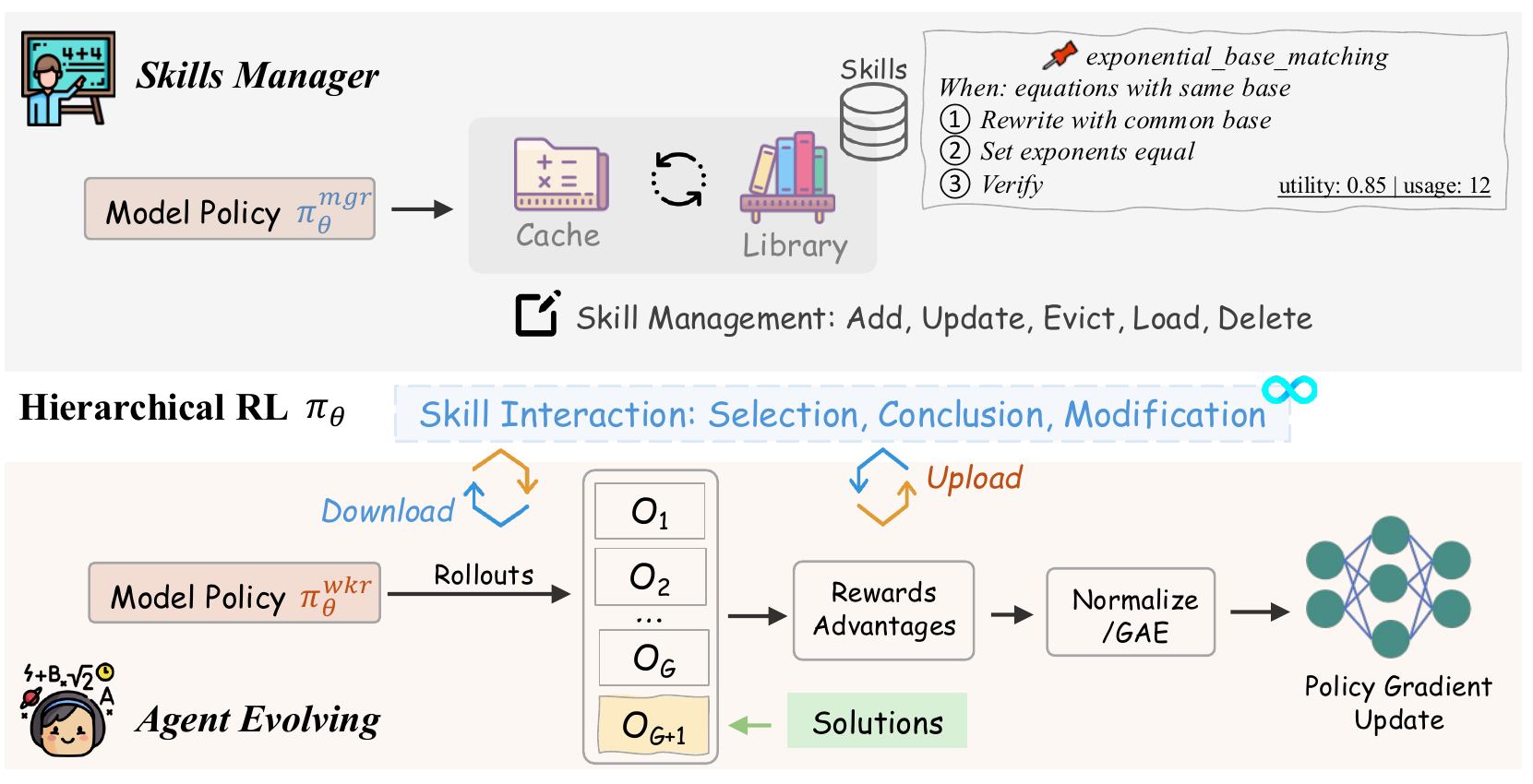}
    \caption{Overview of \n. The shared policy $\pi_\theta$ operates as both Skills Manager and Worker. Before each rollout, the manager scores cache entries via conditional log-probability and injects the selected skill into the prompt (\textit{Download}). After reward computation, an additional rollout $O_{G+1}$ distills successful solutions into a structured skill document (\textit{Upload}). The two-tier library consists of a compact cache (active pool for selection) and a larger reservoir (archive for future promotion), maintained by five operations: \textsc{Add}, \textsc{Update}, \textsc{Evict}, \textsc{Load}, and \textsc{Delete}.}
    \label{fig:overview}
\end{figure}

As illustrated in Figure~\ref{fig:overview}, \n\ resolves both issues through a unified Manager-Worker architecture in which a single policy $\pi_\theta$ governs all skill interactions. 
Through a Download channel, the manager retrieves a relevant skill to condition the worker's rollouts and through an Upload channel, the manager distills successful solution traces into new skill documents. Five management operations (\textsc{Add}, \textsc{Update}, \textsc{Evict}, \textsc{Load}, \textsc{Delete}) maintain library quality throughout training, ensuring that the library co-evolves with the policy under a shared RL objective. 
We formalize this coupling as an evolving-skill MDP with a hierarchical policy (\S\ref{sec:formulation}), then describe the two-phase training framework (\S\ref{sec:training}).

\subsection{Evolving-Skill MDP and Hierarchical Policy}
\label{sec:formulation}

Skill library agents~\citep{wang2023voyager,nguyen2024dynasaur} have shown 
that equipping agents with reusable, structured skills improves their 
ability to handle specialized tasks.
Bringing this paradigm into mathematical reasoning under RL training 
introduces a fundamental challenge: the library is not given a priori 
but must be constructed and refined by the agent itself, coupling library 
dynamics with policy optimization.

To formalize this coupling, we define an Evolving-Skill MDP (ES-MDP) as 
the tuple $\langle \mathcal{Q}, \mathcal{M}, \mathcal{A}, \mathcal{T}, R 
\rangle$, where $\mathcal{Q}$ is the query distribution, $\mathcal{M}$ is 
the space of library configurations, $\mathcal{A}$ is the action space 
covering skill management operations, $\mathcal{T}: \mathcal{M} \times 
\mathcal{A} \to \mathcal{M}$ is the library transition function, and 
$R: \mathcal{Q} \times \mathcal{M} \times \mathcal{A} \to \{r_0, r_1, r_2\}$ 
with $r_2 > r_1 > r_0$ is the hierarchical reward.
The augmented state at step $t$ is $s_t = (q_t, M_t)$, where 
$q_t \sim \mathcal{D}$ is sampled exogenously and $M_t$ is the library 
state shaped by the agent's preceding actions. Each library entry 
$e_k = (m_k, u_k) \in M_t$ pairs a skill document $m_k$ with a scalar 
utility estimate $u_k \in \mathbb{R}$ maintained via exponential moving 
average.
Decision-making follows a Manager-Worker hierarchy realized by a shared 
policy $\pi_\theta$. The manager governs skill selection before task 
execution and skill generation after; the worker generates the solution 
trace $\tau_t$. The joint probability over skill management actions $a_t$ 
and solution trace $\tau_t$ factorizes as:
\begin{equation}
  \pi_\theta(a_t, \tau_t \mid q_t, M_t) =
    \pi_\theta^{\mathrm{mgr}}(z_t \mid q_t, M_t)
    \cdot
    \pi_\theta^{\mathrm{wkr}}(\tau_t \mid q_t, \tilde{m}_{z_t})
    \cdot
    \pi_\theta^{\mathrm{lib}}(a_t^{\mathrm{lib}} \mid q_t, M_t, \tau_t)
\label{eq:hier-policy}
\end{equation}
where the three factors correspond to skill selection, solution generation, 
and skill generation with library update, respectively. The component 
$\pi_\theta^{\mathrm{lib}}$ generates a new skill document conditioned on 
the query and positive-advantage traces $\tau^+$ from the preceding 
rollouts. All three components share parameters $\theta$ but operate under 
different conditioning contexts.

The hierarchical policy operates through three interconnected components:
\begin{itemize}[leftmargin=1.5em]

\item \textbf{Skill Selection and Solution Generation.}
Unlike prior approaches that delegate skill retrieval to an external model, \n\ performs selection through $\pi_\theta$ itself. For each candidate skill $m_k \in M_t$, the policy scores query-skill relevance via conditional log-probability:
\begin{equation}
 s_k = \sum_{l=1}^{|m_k|} \log \pi_\theta(m_k^{(l)} \mid q_t,\, m_k^{(<l)})
\label{eq:skill-score}
\end{equation}
The manager converts these scores into a selection distribution and samples from an $\varepsilon$-greedy mixture:
\begin{equation}
  p_k = \frac{\exp(s_k / \sigma)}{\sum_{j} \exp(s_j / \sigma)}, \qquad
  z_t \sim (1 - \varepsilon) \cdot \delta_{{\argmax}_k p_k} + \varepsilon \cdot 
  \mathrm{Uniform}(M_t)
\label{eq:selection}
\end{equation}
where $\sigma$ is a temperature parameter. 
To prevent injection of marginally relevant skills, a confidence gate 
admits the selected skill only when $\max_k p_k \geq \delta$; 
otherwise the worker solves the query unaided. 
When a skill passes the gate, its document is prepended to the input context and the worker generates trace $\tau_t$ conditioned on the augmented prompt. 
Because the same parameters $\theta$ govern both selection and generation, the advantage signal from the hierarchical reward propagates end-to-end through the selection mechanism.

\item \textbf{Skill Generation and Library Management.}
Beyond the $G$ solution rollouts, the manager executes a dedicated skill generation rollout $O_{G+1}$, conditioned on the original query $q$ together with the positive-advantage traces 
$\tau^+ = \{\tau_i \mid \hat{A}_i > 0\}$. 
The grounding in concrete successful solutions turns skill generation from open-ended strategy induction into structured summarization: 
$\pi_\theta$ extracts reasoning patterns from $\tau^+$ into a skill document $m^{\mathrm{new}}$ following a uniform schema comprising skill name, problem type, key insight, step-by-step method, and verification check. 
And the uniform format ensures that log-probability scores in 
Eq.~\ref{eq:skill-score} reflect semantic relevance rather than surface 
variation across skill documents.

The skill library adopts a two-tier architecture: a cache 
$M^{\mathrm{c}}$ with capacity $C_c$ serves as the active pool for 
selection, while a reservoir $M^{\mathrm{r}}$ with capacity $C_r$ stores surplus skills for future promotion, with 
$M_t = M^{\mathrm{c}}_t \cup M^{\mathrm{r}}_t$. 
New skills produced by $O_{G+1}$ enter the cache via \textit{Upload}, 
and selected skills are injected into the worker's prompt via \textit{Download}. 
Five operations maintain library quality: \textsc{Add}, 
\textsc{Update} with $u_k \leftarrow \beta \, u_k + (1-\beta) \, r$, 
\textsc{Evict}, \textsc{Load}, and \textsc{Delete}, collectively governed by:
\begin{equation}
  M^{\mathrm{c}}_{t+1}, M^{\mathrm{r}}_{t+1} = \mathcal{T}\!\bigl(M^{\mathrm{c}}_t, 
  M^{\mathrm{r}}_t, m^{\mathrm{new}}\bigr), \quad 
  m^{\mathrm{new}} \sim \pi_\theta(\cdot \mid q_t, \tau^+)
\label{eq:lib-transition}
\end{equation}

\item \textbf{Hierarchical Reward.}
The reward combines a task completion signal $r^{\mathrm{task}} \in \{0,1\}$ with a skill utilization bonus $r^{\mathrm{skill}} \geq 0$, granted only when the agent both solves the task and uses a selected skill. 
The composite reward $R = r^{\mathrm{task}} + r^{\mathrm{skill}} 
\in \{r_0, r_1, r_2\}$ with $r_2 > r_1 > r_0$ distinguishes correct 
solutions with skill use ($r_2$), without ($r_1$), and incorrect 
solutions ($r_0$). Within a rollout group containing both $r_2$ and 
$r_1$ trajectories, group-relative advantage assigns strictly higher 
values to the skill-augmented ones for any reward structure satisfying 
$r_2 > r_1 > r_0$; we set $r_0{=}0,\, r_1{=}1,\, r_2{=}2$ in 
experiments. As the policy learns to leverage skills more effectively, 
$O_{G+1}$ produces higher-quality documents from stronger solution 
traces, creating a co-evolutionary dynamic between policy improvement and library enrichment.
\end{itemize}

\begin{algorithm}[!t]
\caption{ARISE: Agent Reasoning with Intrinsic Skill Evolution}
\label{alg:main}
\begin{algorithmic}[1]
\REQUIRE Dataset $\mathcal{D}$, policy $\pi_{\theta_0}$, group size $G$, 
warm-up steps $N_w$, seed skills $M_{\mathrm{seed}}$
\STATE $M^{\mathrm{c}} \leftarrow M_{\mathrm{seed}}$, \; $M^{\mathrm{r}} \leftarrow \emptyset$
\STATE \phasecomment{Phase I: Build base policy and populate skill library}
\FOR{$t = 1, \ldots, N_w$}
  \FOR{each $q \sim \mathcal{D}$}
    \STATE Sample $G$ trajectories $\tau_i \sim \pi_{\theta_{\mathrm{old}}}(\cdot \mid q)$; compute $r_i^{\mathrm{task}}$ and $\hat{A}_i$ (Eq.~\ref{eq:advantage})
    \STATE Collect $\tau^+$; generate $m^{\mathrm{new}} \sim \pi_\theta(\cdot \mid q, \tau^+)$; update $M^{\mathrm{c}}, M^{\mathrm{r}}$ (Eq.~\ref{eq:lib-transition})
    \sidecomment{skill generation $O_{G+1}$}
  \ENDFOR
  \STATE Update $\theta$ via $\mathcal{L}_{\mathrm{GRPO}}(\theta)$ (Eq.~\ref{eq:grpo})
\ENDFOR
\STATE \phasecomment{Phase II: Activate full hierarchical pipeline}
\FOR{$t = N_w + 1, N_w + 2, \ldots$}
  \FOR{each $q \sim \mathcal{D}$}
    \STATE Score and select skill $z_i$ via Eq.~\ref{eq:skill-score},~\ref{eq:selection}
    \sidecomment{manager: selection}
    \STATE Sample $G$ trajectories $\tau_i \sim \pi_{\theta_{\mathrm{old}}}(\cdot \mid q, \tilde{m}_{z_i})$
    \sidecomment{worker: conditioned rollouts}
    \STATE Compute $R_i \in \{0,1,2\}$ and $\hat{A}_i$; update $u_{z_i} \leftarrow \beta \, u_{z_i} + (1-\beta)\, R_i$
    \STATE Collect $\tau^+$; generate $m^{\mathrm{new}} \sim \pi_\theta(\cdot \mid q, \tau^+)$; update $M^{\mathrm{c}}, M^{\mathrm{r}}$ (Eq.~\ref{eq:lib-transition})
    \sidecomment{skill generation $O_{G+1}$}
  \ENDFOR
  \STATE Update $\theta$ via $\mathcal{L}_{\mathrm{GRPO}}(\theta)$ (Eq.~\ref{eq:grpo})
\ENDFOR
\end{algorithmic}
\end{algorithm}

\subsection{Two-Phase Training}
\label{sec:training}
Training proceeds in two phases, summarized in Algorithm~\ref{alg:main}. 
In Phase~I, the policy is warmed up with standard GRPO on binary task rewards while the skill library is silently populated from successful traces. 
In Phase~II, the full hierarchical pipeline activates: the manager begins 
selecting skills, the reward switches from $r^{\mathrm{task}} \in \{0,1\}$ to $R \in \{0,1,2\}$, and policy optimization, skill selection, and library enrichment proceed jointly.

\textbf{Phase~I: Warm-Up.}
The library is initialized with a small set of seed skills encoding generic mathematical reasoning heuristics (e.g., ``extract key quantities,'' ``map counting problems to structured objects''), following the same schema as generated skills. 
The library is initialized with a small set of seed skills encoding generic mathematical reasoning heuristics (e.g., ``extract key quantities,'' ``map counting problems to structured objects''), following the same schema as generated skills. 
During the first $N_w$ steps, skill selection is disabled and the policy 
is trained with standard GRPO on the binary reward $r^{\mathrm{task}}$. 
Group-relative advantages over $G$ rollout trajectories per query are:
\begin{equation}
  \hat{A}_i = \frac{r_i - \mu_G}{\sigma_G + \epsilon}, \qquad 
  \mu_G = \frac{1}{G}\sum_{i=1}^{G} r_i, \quad 
  \sigma_G = \sqrt{\frac{1}{G}\sum_{i=1}^{G}(r_i - \mu_G)^2}
\label{eq:advantage}
\end{equation}
where $r_i$ denotes $r_i^{\mathrm{task}}$ in Phase~I and $R_i$ in Phase~II. 
The policy is updated via the clipped surrogate objective:
\begin{equation}
  \mathcal{L}_{\mathrm{GRPO}}(\theta) = \frac{1}{G} \sum_{i=1}^{G} 
  \frac{1}{|\tau_i|} 
  \sum_{l=1}^{|\tau_i|} \min\!\bigl( \rho_{i,l}\, \hat{A}_i,\; 
  \mathrm{clip}(\rho_{i,l}, 1{-}\epsilon_c, 1{+}\epsilon_c)\, \hat{A}_i \bigr)
\label{eq:grpo}
\end{equation}
where $\rho_{i,l} = \pi_\theta(x_{i,l} \mid q, x_{i,<l})\,/\,
\pi_{\theta_{\mathrm{old}}}(x_{i,l} \mid q, x_{i,<l})$ is the per-token 
importance sampling ratio and $\epsilon_c$ is the clipping parameter. 
While skill selection remains inactive during warm-up, the skill generation rollout $O_{G+1}$ executes at every step, summarizing positive-advantage traces $\tau^+ = \{\tau_i \mid \hat{A}_i > 0\}$ into structured documents via Eq.~\ref{eq:lib-transition}. 

\textbf{Phase~II: Skill-Augmented GRPO.}
Starting from step $N_w{+}1$, the manager scores all cache entries via 
Eq.~\ref{eq:skill-score}, selects a skill through the $\varepsilon$-greedy mechanism of Eq.~\ref{eq:selection}, and the worker generates $G$ solutions conditioned on the augmented prompt. The hierarchical reward $R_i \in \{0,1,2\}$ replaces $r_i^{\mathrm{task}}$ in the advantage calculation of Eq.~\ref{eq:advantage}, and the importance sampling ratio now reflects skill-conditioned generation: 
$\rho_{i,l} = \pi_\theta(x_{i,l} \mid q_t, \tilde{m}_{z_i}, x_{i,<l})\,/\,
\pi_{\theta_{\mathrm{old}}}(x_{i,l} \mid q_t, \tilde{m}_{z_i}, x_{i,<l})$.

The shift from $r^{\mathrm{task}} \in \{0,1\}$ to $R \in \{0,1,2\}$ 
directly shapes the policy gradient. Within a rollout group containing 
trajectories that solve the problem both with ($R{=}2$) and without 
($R{=}1$) skill augmentation, the group-relative advantage assigns positive values to skill-augmented trajectories and negative values to unaugmented ones, even though both are correct. 
\section{Results}
\label{sec:results}

\subsection{Implementation Details}
\label{sec:impl}

We train all methods on the DeepScaleR dataset~\citep{deepscaler2025}, comprising approximately 40K problem-answer pairs from AMC, AIME, MATH, and OlympiadBench, using two instruction-tuned base models: Qwen3-4B-Instruct-2507~\citep{qwen3technicalreport} and Phi-4-mini-instruct~\citep{abouelenin2025phi}. 
All methods use GRPO with group size $G{=}8$ under the same computational budget.
We compare against three categories of baselines. \textit{GRPO Family (Vanilla)} includes GRPO~\citep{guo2025deepseek}, Dr.GRPO~\citep{liu2025understanding}, DAPO~\citep{yu2025dapo}, and GSPO~\citep{zheng2025group}, representing policy optimization without external knowledge. \textit{Memory and Skill-Augmented} methods include EvolveR~\citep{wu2025evolver}, SimpleMem~\citep{liu2026simplemem}, and 
SkillRL~\citep{xia2026skillrl}, each integrated with GRPO and adapted to our mathematical reasoning setting.
We evaluate on two benchmark groups. 
In-distribution competition benchmarks 
(AMC 2023\footnote{\url{https://huggingface.co/datasets/AI-MO/aimo-validation-amc}}, AIME 2024\&2025\footnote{\url{https://huggingface.co/datasets/AI-MO/aimo-validation-aime}}) share the same problem type as the training set but have no temporal overlap with it. Out-of-distribution Omni-MATH~\citep{gao2024omni}, comprising 4,428 Olympiad-level problems across Algebra, Number Theory, Combinatorics, and Geometry, assesses generalization beyond the training distribution. 
All results report average Pass@1 over 32 runs.

\subsection{Performance}

\begin{table*}[!t]
\centering
\caption{\textbf{Main results on mathematical reasoning benchmarks.} We report average Pass@1 accuracy (\%) over 32 independent runs. All methods use $G{=}8$ and are trained on the DeepScaleR dataset. \textbf{Bold} indicates the best per column within each base model. \colorbox{rowgreen}{Shaded rows} represent our proposed method, with subscripts indicating absolute change relative to the GRPO baseline.}
\label{tab:main}
\small
\setlength{\tabcolsep}{4pt}
\resizebox{\textwidth}{!}{%
\begin{tabular}{@{}l *{3}{c} @{\hspace{8pt}} *{4}{>{\centering\arraybackslash}p{1.5cm}} c @{}}
\toprule
 & \multicolumn{3}{c}{\textbf{In-Distribution} \, \small\textit{Competition}}
 & \multicolumn{5}{c}{\textbf{Out-of-Distribution} \, \small\textit{Omni-MATH}} \\
\cmidrule(r){2-4} \cmidrule(l){5-9}
\textbf{Method} & AMC 23 & AIME 24 & AIME 25
 & Algebra & Num.Thy & Combin. & Geometry & Avg.$^\dag$ \\
\midrule
\addlinespace[3pt]
\multicolumn{9}{@{}l}{\textbf{Qwen3-4B-Instruct-2507}} \\
\addlinespace[2pt]
Baseline & 69.6 & 53.6 & 45.6 & 30.1 & 21.2 & 13.0 & 20.5 & 21.2 \\
\midrule
\addlinespace[2pt]
\multicolumn{9}{@{}l}{\textit{GRPO Family (Vanilla)}} \\
\addlinespace[2pt]
GRPO~\citep{guo2025deepseek} & 72.9 & 54.1 & 46.5 & 33.5 & 24.3 & 15.1 & 22.8 & 23.9 \\
Dr.GRPO~\citep{liu2025understanding} & 72.8 & 54.3 & 46.6 & 33.8 & 24.1 & 15.3 & 22.6 & 23.9 \\
DAPO~\citep{yu2025dapo} & 74.2 & 55.3 & 47.6 & 35.2 & 25.7 & 16.3 & 23.9 & 25.3 \\
GSPO~\citep{zheng2025group} & 73.8 & 54.9 & \textbf{48.5} & 34.8 & 25.4 & 16.1 & \textbf{25.8} & 25.5 \\
\midrule
\addlinespace[2pt]
\multicolumn{9}{@{}l}{\textit{Memory-Augmented}} \\
\addlinespace[2pt]
EvolveR + GRPO~\citep{wu2025evolver}    & 73.6 & 54.8 & 46.3 & 34.5 & 25.1 & 15.4 & 23.5 & 24.6 \\
SimpleMem + GRPO~\citep{liu2026simplemem}  & 73.1 & 54.5 & 46.9 & 33.9 & 24.6 & 16.5 & 23.1 & 24.5 \\
\midrule
\addlinespace[2pt]
\oursrow \textbf{\n~+ GRPO}
  & \textbf{75.4}\inc{2.5}
  & \textbf{56.4}\inc{2.3}
  & 48.3\inc{1.8}
  & \textbf{37.0}\inc{3.5}
  & \textbf{27.2}\inc{2.9}
  & \textbf{17.6}\inc{2.5}
  & 25.5\inc{2.7}
  & \textbf{26.8}\inc{2.9} \\
\specialrule{\heavyrulewidth}{3pt}{3pt}
\addlinespace[3pt]
\multicolumn{9}{@{}l}{\textbf{Phi-4-mini-instruct-3.8B}} \\
\addlinespace[2pt]
Baseline & 36.3 & 9.7 & 4.3 & 10.3 & 7.1 & 4.4 & 6.5 & 7.1 \\
\midrule
\addlinespace[2pt]
\multicolumn{9}{@{}l}{\textit{GRPO Family (Vanilla)}} \\
\addlinespace[2pt]
GRPO~\citep{guo2025deepseek} & 42.7 & 10.3 & 4.7 & 13.2 & 9.5 & 6.0 & 8.8 & 9.4 \\
Dr.GRPO~\citep{liu2025understanding} & 42.5 & 10.6 & 4.6 & 13.4 & 9.3 & 6.1 & 8.6 & 9.4 \\
DAPO~\citep{yu2025dapo} & 44.1 & 11.5 & \textbf{6.1} & 14.6 & 10.7 & 7.0 & 9.5 & 10.5 \\
GSPO~\citep{zheng2025group} & 43.6 & 11.8 & 5.0 & 14.2 & 10.3 & 6.7 & 9.8 & 10.2 \\
\midrule
\addlinespace[2pt]
\multicolumn{9}{@{}l}{\textit{Memory-Augmented}} \\
\addlinespace[2pt]
EvolveR + GRPO~\citep{wu2025evolver}    & 43.2 & 10.9 & 5.0 & 14.0 & 10.1 & 6.2 & 9.4 & 9.9 \\
SimpleMem + GRPO~\citep{liu2026simplemem}  & 43.0 & 10.7 & 4.9 & 13.6 & 9.8 & 6.4 & 9.0 & 9.7 \\
\midrule
\addlinespace[2pt]
\oursrow \textbf{\n~+ GRPO}
  & \textbf{45.3}\inc{2.6}
  & \textbf{12.6}\inc{2.3}
  & 5.9\inc{1.2}
  & \textbf{15.9}\inc{2.7}
  & \textbf{11.5}\inc{2.0}
  & \textbf{7.7}\inc{1.7}
  & \textbf{10.6}\inc{1.8}
  & \textbf{11.4}\inc{2.0} \\
\bottomrule
\end{tabular}%
}
\vspace{2pt}
{\textbf{Note:} \footnotesize $^\dag$\,Macro average (arithmetic mean) over four Omni-MATH domains.
In-distribution benchmarks share problem types with training data but have no temporal overlap. Omni-MATH sub-categories follow the official domain taxonomy.}
\end{table*}

Table~\ref{tab:main} summarizes the main results across both base models and all benchmark groups.
On Qwen3-4B-Instruct-2507, \n\ surpasses all baselines across every 
benchmark, outperforming the strongest GRPO-family methods by over one point on in-distribution tasks. On Phi-4-mini-instruct, where the base model has substantially weaker mathematical ability, \n\ still achieves the highest scores across all benchmarks, confirming that the hierarchical skill mechanism remains effective even when positive training signals are sparse.

The advantage of \n\ is more pronounced on out-of-distribution evaluation. On Omni-MATH, \n\ improves over GRPO by 2.9 and 1.9 points in average accuracy on the two base models respectively, with gains observed across all four sub-domains. The largest improvements appear in Algebra, where the skill library accumulates the most reusable reasoning patterns. This suggests that the evolving skill library facilitates transfer to unseen mathematical domains rather than merely reinforcing patterns present in the training set.

Memory-augmented methods like EvolveR and SimpleMem, improve upon GRPO but do not consistently surpass DAPO or GSPO, occasionally matching them on individual benchmarks while falling short on others. 
The result reflects their reliance on external retrievers decoupled from the policy gradient. In contrast, \n\ achieves substantially larger gains through end-to-end policy-driven skill selection, where the advantage signal directly shapes which skills are retrieved and retained.

\subsection{Ablation and Analysis}

\begin{wrapfigure}{r}{0.48\linewidth}
\vspace{-10pt}
\centering
\captionof{table}{Ablation on Qwen3-4B with Pass@1 (\%).}
\label{tab:ablation}
\small
\setlength{\tabcolsep}{3pt}
\begin{tabular}{@{}cl ccc@{}}
\toprule
 & \textbf{Variant} & AIME 24 & AIME 25 & Omni \\
\midrule
 & \n\ (full)        & \textbf{56.4} & \textbf{49.1} & \textbf{28.0} \\
\midrule
(a) & Binary reward     & 54.7 & 47.1 & 25.7 \\
(b) & Random select.    & 55.1 & 47.8 & 26.3 \\
(c) & w/o $O_{G+1}$     & 55.0 & 47.5 & 26.0 \\
(d) & w/o conf. gate    & 56.0 & 48.7 & 27.4 \\
\bottomrule
\end{tabular}
\vspace{4pt}
\includegraphics[width=\linewidth]{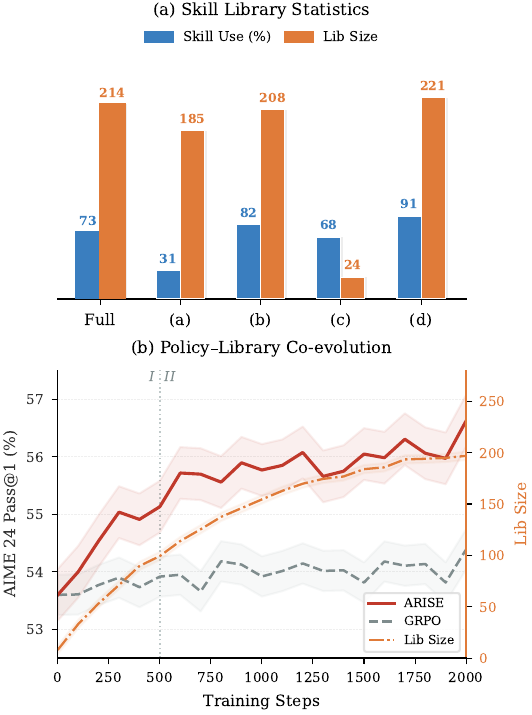}
\captionof{figure}{(a) Skill library statistics. (b) Policy library co-evolution.}
\label{fig:ablation}
\vspace{-10pt}
\end{wrapfigure}

We ablate key design choices of \n\ on Qwen3-4B-Instruct-2507 ($G{=}8$) and report both accuracy and skill library behavior. As shown in Ablation Table, using binary reward causes the largest accuracy drop and reduces skill utilization from 73\% to 31\%, confirming that the hierarchical signal is the primary driver of skill adoption. 
Random skill injection maintains high utilization by construction, yet accuracy degrades because mismatched 
skills provide irrelevant context. 
Removing the skill generation rollout $O_{G+1}$ freezes the library at 24 seed skills, with the largest impact on Omni-MATH where static heuristics cannot cover Olympiad-level diversity. Removing the confidence gate has the smallest effect but pushes utilization to 91\%, confirming its role as a noise filter for borderline cases.

As shown in Figure~\ref{fig:ablation}(b), both \n\ and GRPO start from 
the same baseline. Once Phase~II activates, \n\ diverges as the library 
grows and useful skills are increasingly leveraged. The accuracy gap 
widens in tandem with library growth, providing direct evidence that 
policy improvement and library enrichment are mutually reinforcing. 
Notably, library size saturates in late Phase~II while accuracy continues 
to improve, suggesting that later-stage gains come from the policy 
learning to select existing skills more effectively rather than from 
accumulating new ones.

\vspace{4pt}
\par\noindent\hspace{0pt}

\section{Conclusion}
\vspace{-2pt}
We presented \n, a hierarchical reinforcement learning framework that 
unifies skill selection, generation, and policy optimization under a 
single shared policy, enabling the skill library to co-evolve with the 
agent throughout training. Experiments on two base models across seven benchmarks demonstrate consistent improvements over GRPO-family baselines and memory-augmented methods. The ablation analysis reveals that the hierarchical reward is the primary driver of skill adoption, and that late-stage accuracy gains stem from improved selection over existing skills rather than continued library expansion, suggesting that library curation matters more than library size once a sufficient skill repertoire is established.
The current framework is evaluated exclusively on mathematical reasoning; extending ARISE to multi-tool agent tasks and code generation, where reusable skills take the form of executable procedures rather than reasoning heuristics, is a promising direction for future work.

\bibliography{ref}
\bibliographystyle{colm2026}

\clearpage
\appendix
\section*{Appendix}

\section{Implementation Details}
\label{app:implementation}

\subsection{Training Configuration and Hyperparameters}
\label{app:training_config}

All experiments are trained on the DeepScaleR dataset~\citep{deepscaler2025}, comprising approximately 40K problem-answer pairs from AMC, AIME, MATH, and OlympiadBench, for 2{,}000 optimization steps.
Training is conducted on $\times$A100-80GB GPUs using the GRPO objective (Eq.~\ref{eq:grpo}) with group size $G{=}8$.
We set the per-device batch size to 4 with gradient accumulation over 2 steps, yielding an effective batch of 64 queries per policy update.
The optimizer is AdamW with learning rate $1\!\times\!10^{-6}$, weight decay $0.01$, and a cosine learning-rate schedule with 50 linear warm-up steps.
The maximum generation length is 4{,}096 tokens for both solution rollouts $O_1,\ldots,O_G$ and the skill generation rollout $O_{G+1}$.
The clipping parameter $\epsilon_c$ in Eq.~\ref{eq:grpo} is $0.2$ for all methods.

Phase~I spans the first $N_w{=}500$ steps with binary reward $r^{\mathrm{task}} \in \{0,1\}$.
Phase~II activates at step 501 with the hierarchical reward $R \in \{0,1,2\}$ and full skill selection.
Following DAPO~\citep{yu2025dapo}, we apply dynamic sampling to filter out zero-variance groups---those where all $G$ trajectories receive the same reward---before advantage estimation, which stabilizes the gradient signal in both phases.

Table~\ref{tab:hyperparams} consolidates all ARISE-specific hyperparameters.

\begin{table}[h]
\centering
\caption{ARISE-specific hyperparameters. Parameters above the double rule are introduced by our method; those below are inherited from the GRPO training infrastructure.}
\label{tab:hyperparams}
\small
\begin{tabular}{lcc}
\toprule
\textbf{Hyperparameter} & \textbf{Symbol} & \textbf{Value} \\
\midrule
\multicolumn{3}{l}{\textit{Skill Selection (Manager)}} \\
Softmax temperature & $\sigma$ & 1.0 \\
$\varepsilon$-greedy exploration rate & $\varepsilon$ & 0.1 \\
Confidence gate threshold & $\delta$ & 0.35 \\
Max skill tokens for scoring & --- & 128 \\
\midrule
\multicolumn{3}{l}{\textit{Skill Library}} \\
Cache capacity & $C_c$ & 10 \\
Reservoir capacity & $C_r$ & 100 \\
Utility EMA coefficient & $\beta$ & 0.9 \\
Number of seed skills & $|M_{\mathrm{seed}}|$ & 5 \\
\midrule
\multicolumn{3}{l}{\textit{Skill Generation ($O_{G+1}$)}} \\
Generation temperature & --- & 0.7 \\
Top-$p$ sampling & --- & 0.95 \\
Max new tokens & --- & 192 \\
Max skill document length & --- & 220 chars \\
\midrule
\multicolumn{3}{l}{\textit{Training Schedule}} \\
Warm-up steps (Phase~I) & $N_w$ & 500 \\
Skill utilization bonus & $r^{\mathrm{skill}}$ & 1.0 \\
\midrule
\midrule
\multicolumn{3}{l}{\textit{Shared GRPO Infrastructure}} \\
Group size & $G$ & 8 \\
Clipping parameter & $\epsilon_c$ & 0.2 \\
Learning rate & --- & $1\!\times\!10^{-6}$ \\
Weight decay & --- & 0.01 \\
Max generation length & --- & 4{,}096 tokens \\
\bottomrule
\end{tabular}
\end{table}

\subsection{Complete Training Pipeline}
\label{app:training_pipeline}

Each training step executes the following sequence, applicable to both phases with the differences noted inline:

\begin{enumerate}[leftmargin=1.8em, itemsep=2pt]
    \item \textbf{Batch loading.} Sample a mini-batch of queries $\{q\} \sim \mathcal{D}$.
    \item \textbf{Skill selection (Phase~II only).} For each query, the manager scores all cache entries via Eq.~\ref{eq:skill-score}, converts scores to a selection distribution (Eq.~\ref{eq:selection}), and samples a skill $z_t$. If $\max_k p_k < \delta$, no skill is injected.
    \item \textbf{Prompt construction.} Prepend the selected skill to the query following the injection format described in Appendix~\ref{app:selection_details}. When no skill is selected, the query is presented without modification.
    \item \textbf{Solution rollouts.} Generate $G$ trajectories $\tau_1,\ldots,\tau_G$ from the current policy $\pi_{\theta_{\mathrm{old}}}$.
    \item \textbf{Reward computation.} Assign $R_i = r^{\mathrm{task}}_i + r^{\mathrm{skill}}_i$ per trajectory (see Appendix~\ref{app:reward_details}). In Phase~I, $r^{\mathrm{skill}}_i=0$ for all $i$.
    \item \textbf{Dynamic sampling.} Remove groups with zero reward variance before advantage estimation, following DAPO~\citep{yu2025dapo}.
    \item \textbf{Advantage estimation and policy update.} Compute group-relative advantages $\hat{A}_i$ via Eq.~\ref{eq:advantage}, compute log-probabilities under $\pi_\theta$ and $\pi_{\theta_{\mathrm{old}}}$, and update $\theta$ via $\mathcal{L}_{\mathrm{GRPO}}(\theta)$ (Eq.~\ref{eq:grpo}).
    \item \textbf{Skill generation ($O_{G+1}$).} Collect positive-advantage traces $\tau^+ = \{\tau_i \mid \hat{A}_i > 0\}$. If $\tau^+ \neq \emptyset$, execute the summary rollout (Appendix~\ref{app:summary_rollout}) to produce a new skill $m_{\mathrm{new}}$.
    \item \textbf{Library maintenance.} Update the utility of the selected skill via EMA; insert $m_{\mathrm{new}}$ into the cache; execute \textsc{Evict}/\textsc{Load}/\textsc{Delete} operations as needed (Appendix~\ref{app:two_tier}).
\end{enumerate}

Steps 2--3 are skipped in Phase~I, and the reward in step~5 reduces to the binary signal $r^{\mathrm{task}}$. Steps 8--9 execute in both phases, ensuring the library is populated before Phase~II begins.

\subsection{Evaluation Protocol}
\label{app:eval_protocol}

All results report average Pass@1 accuracy over 32 independent evaluation runs. In each run the model generates a single solution per problem with greedy decoding (temperature $0$, top-$p\,{=}\,1.0$).
At evaluation time the full ARISE pipeline is active: the manager selects a skill from the cache via Eq.~\ref{eq:skill-score}--\ref{eq:selection} with $\varepsilon{=}0$ (exploration disabled), and the worker generates the solution conditioned on the augmented prompt. If no skill exceeds the confidence threshold $\delta$, the model solves the problem unaided.

For in-distribution benchmarks (AMC~2023, AIME~2024\&2025), answers are verified by exact numerical match. For Omni-MATH, we adopt the official evaluation protocol of~\citet{gao2024omni}, which uses symbolic equivalence checking across its four domains.
To ensure fair comparison, all baselines and ablation variants are evaluated with the same decoding configuration and the same 32-run averaging procedure.

\section{Formal Properties of the Evolving-Skill MDP}
\label{app:formulation}

This section provides additional formal details on the ES-MDP introduced in Section~\ref{sec:formulation}.

\subsection{State Space and Transition Structure}
\label{app:state_space}

The augmented state $s_t = (q_t, M_t)$ decomposes into an exogenous component $q_t \sim \mathcal{D}$ (the query, sampled independently at each step) and an endogenous component $M_t$ (the skill library, shaped by the agent's preceding actions). This decomposition distinguishes the ES-MDP from a standard MDP in two respects.

First, the library transition $\mathcal{T}$ is deterministic conditioned on the library management action $a_t^{\mathrm{lib}}$: given the current library $M_t$ and a validated skill document $m^{\mathrm{new}}$, the five management operations (Section~\ref{sec:formulation}) produce a unique successor $M_{t+1}$. Stochasticity enters only through the policy's generation of $m^{\mathrm{new}}$ via $O_{G+1}$, not through the transition itself.

Second, the library state $M_t$ grows monotonically in information: new skills can be added, existing skills can be updated or relocated between tiers, and only reservoir entries meeting strict low-utility criteria are deleted. This monotonicity provides an implicit curriculum---later training steps operate with a richer library than earlier ones, enabling the policy to condition on increasingly specialized reasoning strategies.

\subsection{Factorization of the Hierarchical Policy}
\label{app:factorization}

The joint action probability in Eq.~\ref{eq:hier-policy} factorizes into three sequential decisions. We expand each factor below.

\paragraph{Manager: Skill Selection.} The manager selects a skill index $z_t$ from the cache $M_t^c$ based on query-skill relevance scores. The conditional distribution is:
\begin{equation}
\pi_\theta^{\mathrm{mgr}}(z_t = k \mid q_t, M_t) = 
\begin{cases}
(1-\varepsilon)\,\mathbb{I}[k = \argmax_j p_j] + \frac{\varepsilon}{|M_t^c|} & \text{if } \max_j p_j \geq \delta \\
\mathbb{I}[k = \varnothing] & \text{otherwise}
\end{cases}
\label{eq:mgr-full}
\end{equation}
where $p_j = \exp(s_j/\sigma) / \sum_{j'}\exp(s_{j'}/\sigma)$ and $z_t = \varnothing$ denotes the no-skill action. The confidence gate threshold $\delta$ ensures that the manager abstains when no candidate is sufficiently relevant, preventing the injection of misleading context.

\paragraph{Worker: Solution Generation.} Conditioned on the selected skill $\tilde{m}_{z_t}$ (or an empty context when $z_t = \varnothing$), the worker generates a solution trace token by token:
\begin{equation}
\pi_\theta^{\mathrm{wkr}}(\tau_t \mid q_t, \tilde{m}_{z_t}) = \prod_{l=1}^{|\tau_t|} \pi_\theta(x_{t,l} \mid q_t, \tilde{m}_{z_t}, x_{t,<l})
\label{eq:wkr-full}
\end{equation}
The skill document, when present, is prepended to the query within the prompt template, so that the worker attends to it through the standard causal attention mechanism without architectural modification.

\paragraph{Manager: Skill Generation.} After observing the rollout outcomes, the manager generates a new skill document from positive-advantage traces:
\begin{equation}
\pi_\theta^{\mathrm{lib}}(a_t^{\mathrm{lib}} \mid q_t, M_t, \tau_t) = \pi_\theta(m^{\mathrm{new}} \mid q_t, \tau^+) \cdot \mathbb{I}[\tau^+ \neq \emptyset]
\label{eq:lib-full}
\end{equation}
When no positive-advantage trace exists ($\tau^+ = \emptyset$), the skill generation rollout is skipped and the library remains unchanged: $M_{t+1} = M_t$.

\subsection{Advantage Estimation under the Hierarchical Reward}
\label{app:advantage_analysis}

The shift from binary reward $r^{\mathrm{task}} \in \{0,1\}$ to the hierarchical reward $R \in \{0,1,2\}$ changes the structure of the group-relative advantage in a precise way. Consider a rollout group of size $G$ containing $n_2$ trajectories with $R{=}2$, $n_1$ with $R{=}1$, and $n_0$ with $R{=}0$, where $n_0 + n_1 + n_2 = G$. The group mean and standard deviation are:
\begin{equation}
\mu_G = \frac{n_1 + 2n_2}{G}, \qquad 
\sigma_G = \sqrt{\frac{n_1(1 - \mu_G)^2 + n_2(2 - \mu_G)^2 + n_0\,\mu_G^2}{G}}
\end{equation}
The advantage for each reward level is then:
\begin{align}
\hat{A}(R{=}2) &= \frac{2 - \mu_G}{\sigma_G + \epsilon} > 0 \\
\hat{A}(R{=}1) &= \frac{1 - \mu_G}{\sigma_G + \epsilon} \quad (\text{sign depends on } \mu_G) \\
\hat{A}(R{=}0) &= \frac{-\mu_G}{\sigma_G + \epsilon} < 0
\end{align}
Crucially, when $\mu_G > 1$ (i.e., when the group contains more $R{=}2$ than $R{=}0$ trajectories), the advantage for $R{=}1$ becomes negative, meaning that correct-but-unaugmented solutions are actively down-weighted relative to skill-augmented ones. This mechanism provides an increasingly strong signal for skill utilization as training progresses and the proportion of successful skill-augmented trajectories grows.

\section{Skill Library Architecture}
\label{app:skill_library}

\subsection{Skill Document Schema and Seed Skills}
\label{app:skill_schema}

Every skill is stored as a JSON document conforming to a uniform five-field schema designed to ensure that log-probability scoring (Eq.~\ref{eq:skill-score}) reflects semantic relevance rather than format variation. Table~\ref{tab:schema} specifies the constraints on each field.

\begin{table}[h]
\centering
\caption{Skill document JSON schema with field-level constraints.}
\label{tab:schema}
\small
\begin{tabular}{lcp{6.8cm}}
\toprule
\textbf{Field} & \textbf{Max Length} & \textbf{Description} \\
\midrule
\texttt{skill\_name}    & 40 chars  & Snake\_case identifier; must be non-empty \\
\texttt{problem\_type}  & ---       & One of: algebra, geometry, combinatorics, number\_theory, calculus, general \\
\texttt{key\_insight}   & 160 chars & Core mathematical idea explaining \emph{why} the approach works \\
\texttt{method}          & 3 steps   & 2--3 procedural steps (each $\leq$100 chars) \\
\texttt{check}           & 100 chars & Verification procedure; defaults to ``Substitute back to verify'' \\
\midrule
\multicolumn{2}{l}{\textit{Total document}} & Hard-capped at 220 characters \\
\bottomrule
\end{tabular}
\end{table}

Figure~\ref{fig:skill_example} shows a representative generated skill.

\begin{figure}[h]
\centering
\begin{tcolorbox}[colback=gray!5, colframe=gray!50, width=0.95\linewidth, boxrule=0.5pt, arc=2pt]
\small
\texttt{\{} \\
\quad \texttt{"skill\_name": "exponential\_base\_matching",} \\
\quad \texttt{"problem\_type": "algebra",} \\
\quad \texttt{"key\_insight": "When both sides of an equation} \\
\quad\quad \texttt{can be expressed as powers of the same base,} \\
\quad\quad \texttt{set exponents equal",} \\
\quad \texttt{"method": [} \\
\quad\quad \texttt{"Rewrite each side with a common base",} \\
\quad\quad \texttt{"Set the exponents equal and solve",} \\
\quad\quad \texttt{"Verify the solution satisfies the original"} \\
\quad \texttt{],} \\
\quad \texttt{"check": "Substitute back into the original equation"} \\
\texttt{\}}
\end{tcolorbox}
\caption{Example of a generated skill document following the uniform schema.}
\label{fig:skill_example}
\end{figure}

The cache is initialized with $|M_{\mathrm{seed}}|{=}5$ seed skills encoding generic mathematical heuristics that are broadly applicable across problem types. Table~\ref{tab:seeds} lists all seed skills and their key insights. Each seed follows the same JSON schema and serves as a warm-start, ensuring that the skill selection mechanism has non-trivial candidates from the very first step of Phase~II.

\begin{table}[h]
\centering
\caption{Seed skills used to initialize the cache at the beginning of training.}
\label{tab:seeds}
\small
\renewcommand{\arraystretch}{1.15}
\begin{tabular}{lp{3.0cm}p{7.0cm}}
\toprule
\textbf{Skill Name} & \textbf{Problem Type} & \textbf{Key Insight} \\
\midrule
\texttt{equation\_setup}         & Algebra     & Translate word-problem quantities into variables and equations before solving \\
\texttt{modular\_arithmetic\_check} & Number Theory & Reduce expressions modulo small primes to constrain or verify integer solutions \\
\texttt{case\_enumeration}       & General     & Systematically split into exhaustive cases and verify each independently \\
\texttt{symmetry\_exploitation}  & General     & Identify and leverage algebraic or geometric symmetry to simplify the problem \\
\texttt{extremal\_principle}     & General     & Consider boundary or extremal configurations to establish bounds or find optima \\
\bottomrule
\end{tabular}
\end{table}

\subsection{Two-Tier Cache-Reservoir Management}
\label{app:two_tier}

The skill library $M_t = M^c_t \cup M^r_t$ consists of a \emph{cache} $M^c$ (capacity $C_c{=}10$) and a \emph{reservoir} $M^r$ (capacity $C_r{=}100$).
The cache is the active pool from which skills are scored and selected at each training step; the reservoir archives surplus skills for potential future promotion.
Each library entry $e_k = (m_k, u_k)$ pairs a skill document $m_k$ with a scalar utility estimate $u_k \in \mathbb{R}$.

Five management operations govern the library dynamics, executed in the order listed below at the end of each training step:

\begin{itemize}[leftmargin=1.5em, itemsep=3pt]
    \item \textbf{\textsc{Update}.} The utility of the skill $z_t$ selected in the current step is updated via exponential moving average: $u_{z_t} \leftarrow \beta \, u_{z_t} + (1-\beta) \, R_t$, where $\beta{=}0.9$ and $R_t$ is the hierarchical reward received by the trajectory. Skills not selected in this step retain their previous utility.
    \item \textbf{\textsc{Add}.} When the summary rollout $O_{G+1}$ produces a valid skill $m_{\mathrm{new}}$ (see Appendix~\ref{app:summary_rollout}), it is inserted into the cache with initial utility $u{=}0$ and usage count~0.
    \item \textbf{\textsc{Evict}.} If the cache size exceeds $C_c$ after insertion, the entry with the lowest utility is moved to the reservoir, preserving its utility and usage metadata.
    \item \textbf{\textsc{Load}.} If any reservoir skill has a utility estimate exceeding that of the lowest-utility cache entry, the two are swapped. This allows previously evicted skills to re-enter the active pool when they become relevant to the current training distribution.
    \item \textbf{\textsc{Delete}.} Reservoir entries satisfying \emph{both} conditions---utility below the 10th percentile of all reservoir utilities, and zero cumulative usage---are permanently removed.
\end{itemize}

The rationale behind the two-tier design is twofold.
First, keeping the cache small ($C_c{=}10$) limits the computational cost of scoring all candidates via Eq.~\ref{eq:skill-score} at each step, while ensuring that only high-utility skills compete for selection.
Second, the reservoir prevents premature discarding of skills that may regain relevance as the training distribution shifts.
The co-evolution evidence in Section~4.3 confirms that library size saturates in late Phase~II while accuracy continues to improve, indicating that the \textsc{Load}/\textsc{Evict} cycle refines the active pool rather than merely expanding it.

\subsection{Execution Order and Edge Cases}
\label{app:edge_cases}

The five operations always execute in the fixed order \textsc{Update} $\to$ \textsc{Add} $\to$ \textsc{Evict} $\to$ \textsc{Load} $\to$ \textsc{Delete}. This ordering ensures that (i)~the utility of the selected skill reflects the most recent reward before any insertion or eviction occurs, (ii)~a newly added skill can immediately trigger an eviction if the cache is full, and (iii)~the \textsc{Load} check runs after eviction so that a just-evicted skill does not immediately re-enter.

Several edge cases arise in practice:
\begin{itemize}[leftmargin=1.5em, itemsep=2pt]
    \item When $O_{G+1}$ fails validation (see Appendix~\ref{app:summary_rollout}), \textsc{Add} is skipped and the remaining operations proceed on the unchanged cache.
    \item When the reservoir is empty, \textsc{Load} and \textsc{Delete} are no-ops.
    \item When the confidence gate rejects all candidates ($z_t = \varnothing$), \textsc{Update} is skipped because no skill was selected, but \textsc{Add} through \textsc{Delete} still execute normally.
    \item When all $G$ trajectories receive the same reward (zero-variance group), dynamic sampling removes the group from advantage estimation, but the skill generation rollout $O_{G+1}$ and library maintenance still proceed if $\tau^+ \neq \emptyset$.
\end{itemize}

\subsection{Policy-Driven Skill Selection and Injection}
\label{app:selection_details}

Skill selection in ARISE is performed by the policy $\pi_\theta$ itself through conditional log-probability scoring, as described in Section~\ref{sec:formulation}. This appendix provides implementation-level details.

For each candidate $m_k \in M^c_t$, the manager computes the score $s_k$ (Eq.~\ref{eq:skill-score}) by feeding the concatenation of query $q_t$ and skill text $m_k$ through $\pi_\theta$ and summing the per-token log-probabilities of $m_k$.
To limit computational cost, skill texts are truncated to 128 tokens during scoring; the full document is used for prompt injection.
With a cache of $C_c{=}10$ skills, this scoring step adds $10 \times 128 = 1{,}280$ extra tokens per query to the forward pass, a modest overhead relative to the 4{,}096-token solution generation.

The selected skill is injected by prepending it to the query with a \texttt{SKILL:} prefix, formatted within the chat template as follows:

\begin{tcolorbox}[colback=gray!5, colframe=gray!50, width=0.95\linewidth, boxrule=0.5pt, arc=2pt]
\small
\texttt{<|im\_start|>user} \\
\texttt{SKILL:\{...\emph{skill JSON}...\}} \\
\texttt{\emph{<math question>}} \\
\texttt{<|im\_end|>} \\
\texttt{<|im\_start|>assistant}
\end{tcolorbox}

The \texttt{SKILL:} prefix acts as a structured delimiter that the model learns to attend to during Phase~II.
When the confidence gate rejects all candidates ($\max_k p_k < \delta$), the query is presented without any prefix, and the trajectory is evaluated under the binary reward $r^{\mathrm{task}} \in \{0,1\}$.

\paragraph{Scoring cost analysis.} 
The log-probability scores are computed via a single batched forward pass over $|M_t^c|$ query-skill concatenations. Since skill documents are capped at 220 characters (approximately 60--80 tokens after tokenization, truncated to 128 for scoring), and the query itself typically spans 50--150 tokens, each scoring pass processes at most $10 \times 278 = 2{,}780$ tokens. On an A100-80GB GPU, this adds approximately 0.3 seconds per query, representing less than 8\% of the total per-step wall time when including the $G{=}8$ solution rollouts.

\section{Skill Generation and Reward}
\label{app:skill_gen_reward}

\subsection{Skill Distillation via Summary Rollout}
\label{app:summary_rollout}

After the $G$ solution rollouts complete, the manager executes a dedicated skill generation rollout $O_{G+1}$, conditioned on the original query $q$ and the positive-advantage traces $\tau^+ = \{\tau_i \mid \hat{A}_i > 0\}$.
This rollout distills the reasoning patterns embodied in $\tau^+$ into a compact, reusable skill document.
Figure~\ref{fig:summary_prompt} shows the prompt template.

\begin{figure}[h]
\centering
\begin{tcolorbox}[colback=blue!3, colframe=blue!40, width=0.95\linewidth, boxrule=0.5pt, arc=2pt, title={\small Summary Rollout Prompt Template}]
\small
\texttt{You are a skill distiller for math reasoning.} \\
\texttt{Given one question and trajectories from the same} \\
\texttt{rollout group, summarize ONE reusable skill.} \\[4pt]
\texttt{Question: \{question\}} \\[2pt]
\texttt{Group trajectories:} \\
\texttt{[SUCCESS \#1] \{clipped\_response\_1 ($\leq$400 chars)\}} \\
\texttt{[SUCCESS \#2] \{clipped\_response\_2 ($\leq$400 chars)\}} \\[4pt]
\texttt{Output MUST be a valid JSON object and nothing else} \\
\texttt{(no markdown/code fences). Schema:} \\
\texttt{\{"skill\_name", "problem\_type", "key\_insight",} \\
\texttt{\ "method", "check"\}} \\[4pt]
\texttt{Rules:} \\
\texttt{- Be generic and transferable; do not copy specific numbers.} \\
\texttt{- Keep the whole skill within 220 characters.} \\
\texttt{- key\_insight is the most important field.} \\
\texttt{- method must contain 2--3 concise steps.} \\
\texttt{- Focus on improving correctness, not style.}
\end{tcolorbox}
\caption{Prompt template used for the skill generation rollout $O_{G+1}$. Up to two successful traces are included, each truncated to 400 characters. The generation uses temperature\,0.7, top-$p$\,0.95, and a maximum of 192 new tokens.}
\label{fig:summary_prompt}
\end{figure}

The generated output undergoes a four-stage validation pipeline before entering the library:
\begin{enumerate}[leftmargin=1.8em, itemsep=1pt]
    \item \textit{Extraction}: isolate the JSON object from the raw generation, stripping any extraneous text or markdown fences;
    \item \textit{Parsing}: deserialize and verify that all five required fields exist and conform to their type constraints (Table~\ref{tab:schema});
    \item \textit{Truncation}: clip any field exceeding its character limit, and discard the document entirely if the total length exceeds 220 characters after clipping;
    \item \textit{Insertion}: add the validated skill to the cache via the \textsc{Add} operation.
\end{enumerate}
If the generation fails to produce valid JSON (e.g., malformed brackets or missing fields), a \emph{direct trace abstraction} fallback constructs a minimal skill from the first successful trace: \texttt{\{"skill\_name":"trace\_abstract", ..., "key\_insight":"Solve by: \{truncated\_response\}"\}}.
In practice, the primary summarization path succeeds in over 85\% of attempts after the warm-up phase; the fallback rate drops further as the policy improves at structured generation during Phase~II.

\paragraph{Design rationale.}
Two choices are worth highlighting.
First, conditioning the summary on concrete successful traces $\tau^+$ turns skill generation from open-ended strategy induction into structured summarization, yielding more precise and transferable skills than unconditioned generation would.
Second, the 220-character hard cap on document length ensures that every skill consumes a comparable number of tokens, so that log-probability scores in Eq.~\ref{eq:skill-score} are dominated by semantic relevance rather than length artifacts. Without this cap, longer documents would accumulate lower (more negative) log-probability sums, biasing selection toward shorter, less informative skills regardless of their content.

\subsection{Hierarchical Reward Details}
\label{app:reward_details}

The hierarchical reward $R = r^{\mathrm{task}} + r^{\mathrm{skill}} \in \{0,1,2\}$ is assigned at the position of the last valid response token in each trajectory. Table~\ref{tab:reward_breakdown} enumerates all four configurations.

\begin{table}[h]
\centering
\caption{Reward assignment by skill usage and task correctness. The bonus $r^{\mathrm{skill}}$ is granted \emph{only} when the agent both uses a selected skill and solves the problem correctly, preventing the policy from being reinforced for skill usage that does not improve reasoning.}
\label{tab:reward_breakdown}
\small
\begin{tabular}{ccccl}
\toprule
\textbf{Skill Used} & \textbf{Task Correct} & $r^{\mathrm{task}}$ & $r^{\mathrm{skill}}$ & $R$ \\
\midrule
No  & No  & 0 & 0 & 0 \quad (incorrect) \\
No  & Yes & 1 & 0 & 1 \quad (correct, unaided) \\
Yes & No  & 0 & 0 & 0 \quad (incorrect despite skill) \\
Yes & Yes & 1 & 1 & 2 \quad (correct with skill) \\
\bottomrule
\end{tabular}
\end{table}

The group-relative advantage (Eq.~\ref{eq:advantage}) computed over $R$ assigns trajectories with $R{=}2$ a strictly higher advantage than those with $R{=}1$ whenever both outcomes coexist in the same group. Critically, incorrect trajectories receive $R{=}0$ regardless of skill usage, so the policy is never reinforced for applying irrelevant skills. The detailed advantage structure under different group compositions is analyzed in Appendix~\ref{app:advantage_analysis}.

\subsection{Information Gain Metric}
\label{app:info_gain}

Beyond the scalar utility maintained per skill (updated via EMA as described in Appendix~\ref{app:two_tier}), we track an \emph{information gain} (IG) metric that measures how much a skill text increases the policy's confidence on known correct solutions:
\begin{equation}
\mathrm{IG}(m_k) = \mathbb{E}_{\tau^+}\!\left[\sum_{t} \log \pi_\theta\!\left(\tau^+_t \mid m_k, q, \tau^+_{<t}\right) - \sum_{t} \log \pi_\theta\!\left(\tau^+_t \mid q, \tau^+_{<t}\right) \right]
\label{eq:info_gain}
\end{equation}
Positive values indicate that conditioning on $m_k$ concentrates probability mass on correct reasoning steps.
Computing Eq.~\ref{eq:info_gain} exactly requires two forward passes per (skill, trace) pair, which is prohibitively expensive at every step.
We therefore maintain a lightweight proxy updated online:
\begin{equation}
\widehat{\mathrm{IG}}(m_k) = \tfrac{1}{2}\!\left(\bar{r}_{m_k} - \bar{r}_{\mathrm{global}}\right) + \tfrac{1}{2}\!\left(\mathrm{SR}_{m_k} - \mathrm{SR}_{\mathrm{global}}\right)
\end{equation}
where $\bar{r}_{m_k}$ and $\mathrm{SR}_{m_k}$ denote the running average reward and success rate for trajectories using skill $m_k$, and $\bar{r}_{\mathrm{global}}$, $\mathrm{SR}_{\mathrm{global}}$ are the corresponding global statistics.
The exact IG (Eq.~\ref{eq:info_gain}) is computed every 100 steps for monitoring and stored alongside each skill's metadata.

The proxy IG serves primarily as a diagnostic tool rather than a training signal. It provides a complementary perspective to the EMA utility: while utility tracks cumulative reward, IG isolates the \emph{causal contribution} of the skill by comparing performance with and without conditioning. Skills with high utility but low or negative IG are likely correlated with easy problems---they are selected when the query happens to be straightforward---rather than causally contributing to the solution. Identifying such skills helps interpret the library's evolution and may inform future work on more selective library management strategies.

\section{Extended Analysis}
\label{app:extended}

\subsection{Hyperparameter Sensitivity}
\label{app:sensitivity}

We study the sensitivity of ARISE to four key hyperparameters on Qwen3-4B-Instruct-2507, evaluating Pass@1 accuracy on AIME~2024 with all other settings fixed at their defaults.
Table~\ref{tab:sensitivity} reports the results.

\begin{table}[h]
\centering
\caption{Hyperparameter sensitivity on AIME 2024 (Qwen3-4B, Pass@1 \%). Default values are \underline{underlined}. Each row varies a single parameter while holding the rest at default.}
\label{tab:sensitivity}
\small
\begin{tabular}{lccccc}
\toprule
\textbf{Confidence gate $\delta$} & 0.15 & 0.25 & \underline{0.35} & 0.45 & 0.55 \\
Accuracy & 55.2 & 55.9 & \textbf{56.4} & 56.1 & 55.0 \\
Skill Use (\%) & 89.3 & 81.6 & 73.0 & 58.2 & 41.7 \\
\midrule
\textbf{Exploration $\varepsilon$} & 0.0 & 0.05 & \underline{0.10} & 0.20 & 0.30 \\
Accuracy & 55.7 & 56.0 & \textbf{56.4} & 55.8 & 54.9 \\
\midrule
\textbf{Cache size $C_c$} & 5 & \underline{10} & 20 & 50 & --- \\
Accuracy & 55.5 & \textbf{56.4} & 56.1 & 55.3 & --- \\
\midrule
\textbf{Warm-up $N_w$} & 200 & 350 & \underline{500} & 750 & 1000 \\
Accuracy & 54.8 & 55.6 & \textbf{56.4} & 56.2 & 55.5 \\
\bottomrule
\end{tabular}
\end{table}

\textit{Confidence gate $\delta$.}
This threshold governs the trade-off between skill utilization and selection precision.
At $\delta{=}0.15$, nearly 90\% of rollouts receive a skill injection, but the liberal threshold admits marginally relevant skills that can introduce misleading context, reducing accuracy by 1.2 points.
Conversely, $\delta{=}0.55$ suppresses skill usage to 42\%, forfeiting the benefit of the hierarchical reward and underperforming the full model by 1.4 points.
The default at $\delta{=}0.35$ injects skills in 73\% of rollouts while maintaining high selection precision.

\textit{Exploration rate $\varepsilon$.}
Performance is robust across $\varepsilon \in [0.05, 0.15]$.
Pure exploitation ($\varepsilon{=}0$) slightly underperforms the default because the manager never evaluates under-explored skills, potentially missing useful entries that recently entered the cache.
Beyond $\varepsilon{=}0.2$, random injection increasingly dominates, degrading accuracy as irrelevant skills corrupt the reasoning context.

\textit{Cache size $C_c$.}
The default of 10 balances diversity against scoring cost.
Halving it to 5 limits the available skill repertoire, while expanding to 50 dilutes the active pool with low-utility entries and increases the per-step scoring overhead by $5\times$ without commensurate accuracy gains.

\textit{Warm-up length $N_w$.}
Shorter warm-ups ($N_w{=}200$) yield the largest accuracy deficit ($-1.6$ points), because Phase~II begins with both a weak base policy and an underpopulated library.
Excessively long warm-ups ($N_w{=}1000$) delay the onset of co-evolution, consuming half the training budget before skill-augmented learning activates.
The default of 500 steps provides a sufficient base policy and a library of approximately 50--80 skills before Phase~II begins.

\subsection{Skill Library Evolution}
\label{app:skill_evolution}

Table~\ref{tab:skill_evolution} illustrates how the skill library evolves qualitatively during training by presenting representative skills at three stages: initialization (seed skills), mid-training (step 1{,}000), and late training (step 2{,}000).

\begin{table}[h]
\centering
\caption{Representative skill documents at different training stages, illustrating the progression from generic heuristics to domain-specific techniques.}
\label{tab:skill_evolution}
\small
\renewcommand{\arraystretch}{1.2}
\begin{tabular}{p{1.2cm}p{2.8cm}p{2.2cm}p{5.5cm}}
\toprule
\textbf{Stage} & \textbf{Skill Name} & \textbf{Type} & \textbf{Key Insight} \\
\midrule
Seed & \texttt{equation\_setup} & Algebra & Translate word-problem quantities into variables and equations before solving \\
Seed & \texttt{case\_enumeration} & General & Systematically split into exhaustive cases and verify each independently \\
\midrule
Step 1k & \texttt{modular\_residue\_analysis} & Num.\ Theory & Reduce expressions modulo $n$ to constrain integer solutions before full computation \\
Step 1k & \texttt{generating\_function\_setup} & Combin.\ & Encode a counting sequence as polynomial coefficients and extract the target term \\
\midrule
Step 2k & \texttt{vieta\_root\_reconstruction} & Algebra & Use Vieta's formulas to express symmetric functions of roots without solving the polynomial \\
Step 2k & \texttt{angle\_chasing\_with\_inscribed} & Geometry & In cyclic quadrilaterals, the inscribed angle theorem converts arc relations to angle equations \\
\bottomrule
\end{tabular}
\end{table}

Several trends emerge.
First, early generated skills (steps 500--800) tend to be broad heuristics resembling the seeds---strategies for setting up equations or checking parity.
As the policy encounters harder problems and the hierarchical reward reinforces successful skill usage, mid-training skills become increasingly specialized to specific domains (Number Theory, Combinatorics).
By late training, the library contains highly targeted techniques (Vieta's formulas, inscribed angle chasing) that transfer effectively to Olympiad-level problems in Omni-MATH.
This specialization trajectory provides a qualitative explanation for the disproportionately large out-of-distribution gains reported in Table~\ref{tab:main}: the evolving library accumulates domain-specific reasoning patterns that the base policy alone does not develop through standard GRPO training.

Second, the co-evolution evidence from Section~4.3 shows that library size plateaus around step 1{,}500 while accuracy continues to rise, confirming that late-stage improvements stem from \emph{better selection over existing skills} rather than from accumulating new ones.
The \textsc{Load}/\textsc{Evict} cycle effectively curates the active cache, promoting specialized skills and retiring those that have become redundant as the policy internalizes their strategies.

\subsection{Phase Transition Dynamics}
\label{app:phase_transition}

The transition from Phase~I to Phase~II at step $N_w{+}1$ introduces two simultaneous changes: the activation of skill selection and the switch from binary to hierarchical reward. To understand their relative contributions, we examine training dynamics around this boundary.

During the first 10--20 steps of Phase~II, skill utilization rises rapidly from 0\% to approximately 40--50\% as the manager begins injecting seed skills and early generated skills. This initial surge is driven primarily by the $\varepsilon$-greedy exploration term rather than confident selection, as the policy has not yet learned to discriminate among candidates. Over the subsequent 100--200 steps, the confidence gate increasingly filters out low-relevance injections and the utilization rate stabilizes around 65--75\%, reflecting genuine query-skill matching.

The reward distribution also shifts notably. In Phase~I, the advantage landscape is binary: trajectories are either correct or incorrect. In Phase~II, the introduction of $R{=}2$ creates a three-level hierarchy that produces more informative gradient signals. Groups containing a mix of $R{=}0$, $R{=}1$, and $R{=}2$ trajectories have higher reward variance, which amplifies the advantage magnitude and provides a stronger learning signal per gradient step compared to the binary case. This partly explains the accelerated accuracy improvement observed in the first 200 steps of Phase~II (Figure~1, middle panel).

\subsection{Token Overhead Analysis}
\label{app:token_overhead}

Prepending skill documents to the prompt incurs additional tokens. Table~\ref{tab:overhead} compares average per-problem token counts on AIME~2024 across methods.

\begin{table}[h]
\centering
\caption{Token overhead comparison on AIME 2024 (Qwen3-4B). Overhead is computed as $(\text{total tokens}_{\text{method}} - \text{total tokens}_{\text{GRPO}}) \,/\, \text{total tokens}_{\text{GRPO}} \times 100\%$, where total tokens = prompt + response.}
\label{tab:overhead}
\small
\begin{tabular}{lcccc}
\toprule
\textbf{Method} & \textbf{Prompt} & \textbf{Response} & \textbf{Total} & \textbf{Overhead} \\
\midrule
GRPO              & 312 & 2{,}847 & 3{,}159 & --- \\
DAPO              & 312 & 2{,}903 & 3{,}215 & +1.8\% \\
GSPO              & 312 & 2{,}871 & 3{,}183 & +0.8\% \\
EvolveR + GRPO    & 489 & 2{,}921 & 3{,}410 & +7.9\% \\
ARISE + GRPO      & 378 & 2{,}876 & 3{,}254 & +3.0\% \\
\bottomrule
\end{tabular}
\end{table}

ARISE adds an average of 66 prompt tokens per problem (the injected skill document), a direct consequence of the 220-character hard cap on skill length.
Despite this modest overhead, ARISE yields a 2.9-point average accuracy gain on Omni-MATH over GRPO (Table~\ref{tab:main}), achieving the best accuracy-to-overhead trade-off among all methods.
By comparison, EvolveR injects longer experience narratives, incurring $7.9\%$ overhead for only a 0.7-point Omni-MATH gain.
This efficiency advantage is visualized in Figure~1 (bottom-right), where ARISE occupies the Pareto-optimal position in the accuracy-gain-versus-overhead space.

DAPO and GSPO have zero prompt overhead (no skill injection) but still show non-zero total overhead due to slightly longer response lengths caused by their modified sampling strategies. The response length of ARISE (2{,}876 tokens) is comparable to vanilla GRPO (2{,}847 tokens), confirming that skill injection does not inflate the solution length---the injected skill guides the reasoning strategy without encouraging more verbose output.

\subsection{Computational Cost Breakdown}
\label{app:compute_cost}

Table~\ref{tab:compute} reports the wall-clock time per training step on 8$\times$A100-80GB GPUs, broken down by component.

\begin{table}[h]
\centering
\caption{Per-step wall-clock time breakdown on 8$\times$A100-80GB (Qwen3-4B, $G{=}8$). Percentages are relative to GRPO baseline.}
\label{tab:compute}
\small
\begin{tabular}{lcc}
\toprule
\textbf{Component} & \textbf{GRPO} & \textbf{ARISE} \\
\midrule
Solution rollouts ($O_1,\ldots,O_G$)     & 38.2s & 39.1s \\
Reward computation                         & 0.4s  & 0.5s  \\
Advantage estimation \& gradient update    & 12.6s & 12.8s \\
Skill scoring \& selection (Phase~II)      & ---   & 2.4s  \\
Skill generation ($O_{G+1}$)               & ---   & 4.8s  \\
Library maintenance                        & ---   & 0.1s  \\
\midrule
\textbf{Total per step}                    & 51.2s & 59.7s \\
\textbf{Overhead vs.\ GRPO}               & ---   & +16.6\% \\
\bottomrule
\end{tabular}
\end{table}

The dominant overhead comes from the skill generation rollout $O_{G+1}$ (4.8s, or 9.4\% of total time), which generates a single sequence of up to 192 tokens. Skill scoring adds 2.4s (4.7\%) due to the $|M_t^c|{=}10$ forward passes for log-probability computation. Library maintenance is negligible ($<$0.2\%). The total per-step overhead of 16.6\% is modest given the consistent accuracy gains observed across all benchmarks, and could be further reduced by batching the scoring passes or caching log-probabilities for skills that remain unchanged between consecutive steps.

\end{document}